\documentclass[11pt,a4paper]{article}
\usepackage[hyperref]{emnlp2019}
\usepackage{times}
\usepackage{latexsym}

\usepackage{amsmath,amsfonts,bm}

\def\eqref#1{equation~\ref{#1}}

\def\1{\bm{1}}

\DeclareMathAlphabet{\mathsfit}{\encodingdefault}{\sfdefault}{m}{sl}
\SetMathAlphabet{\mathsfit}{bold}{\encodingdefault}{\sfdefault}{bx}{n}

\usepackage{hyperref}

\usepackage{url}
\aclfinalcopy
\usepackage{bbold} 
\usepackage[]{todonotes} 
\usepackage{comment} 
\usepackage{bm} 
\usepackage{amssymb} 
\usepackage{tabularx,multirow}
\usepackage[toc,page]{appendix}
\usepackage[symbol]{footmisc}
\usepackage{adjustbox}
\usepackage{color}

\newcommand{\mnno}{\texttt{mNNO}}

\newcolumntype{Y}{>{\centering\arraybackslash}X}
\newcolumntype{C}[1]{>{\centering}p{#1}}
\newcolumntype{O}[1]{>{\raggedleft}p{#1}}

\newcolumntype{R}[2]{%
    >{\adjustbox{angle=#1,lap=\width-(#2)}\bgroup}%
    l%
    <{\egroup}%
}

\newcommand*{\Resize}[2]{\resizebox{#1}{!}{$#2$}}%

\title{Incorporating Visual Semantics into Sentence Representations\\ within a Grounded Space}

\author{Patrick Bordes\thanks{\hspace{0.1cm} Equal contribution.}\hspace{0.1cm}$^1$ \and \textbf{\'Eloi Zablocki}\footnotemark[1]\hspace{0.1cm}$^1$ \and \textbf{Laure Soulier}$^1$ \\ \vspace{0.2cm} \and \textbf{Benjamin Piwowarski}$^1$ \and \textbf{Patrick Gallinari}$^{1,2}$ \\$^1$ Sorbonne Universit\'e, CNRS, LIP6, F-75005 Paris, France \\\vspace{0.1cm}$^2$ Criteo AI Lab, Paris, France\\\texttt{firstname.name@lip6.fr}}

\begin{document}

\maketitle

\begin{abstract}
\textit{Language grounding} is an active field aiming at enriching textual representations with visual information. Generally, textual and visual elements are embedded in the same representation space, which implicitly assumes a one-to-one correspondence between modalities. This hypothesis does not hold when representing words, and becomes problematic when used to learn sentence representations --- the focus of this paper --- as a visual scene can be described by a wide variety of sentences. To overcome this limitation, we propose to transfer visual information to textual representations by learning an intermediate representation space: the grounded space. We further propose two new complementary objectives ensuring that (1) sentences associated with the same visual content are close in the grounded space and (2) similarities between related elements are preserved across modalities. We show that this model outperforms the previous state-of-the-art on classification and semantic relatedness tasks.
\end{abstract}

\section{Introduction}

Representing text by vectors that capture meaningful semantics is a long-standing issue in Artificial Intelligence. Distributional Semantic Models \cite{mikolov,elmo} are well-known recent efforts in this direction, based on the \textit{distributional hypothesis} \cite{harris}. They rely on large text corpora to learn word embeddings. 
At another granularity level, having high-quality and general-purpose sentence representations is crucial for all models that encode sentences into semantic vectors, such as the ones used in machine translation \cite{DBLP:journals/corr/BahdanauCB14} or relation extraction \cite{relationextraction}. 
Moreover, encoding semantics of sentences is paramount because sentences describe relationships between objects, and thus convey complex and high-level knowledge better than individual words \cite{norman1972memory}. 

Relying only on text can lead to biased representations and unrealistic predictions such as ``\textit{the sky is green}'' \cite{baroni}. Besides, it has been shown that human understanding of language is \emph{grounded} in physical reality and perceptual experience \cite{Fincher-Kiefer2001}.
To overcome this limitation, an emerging approach is to \textit{ground} language in the visual world: this consists in leveraging visual information, usually from images, to enrich textual representations.
\footnote{In the Computer Vision community, \textit{grounding} can also refer to the task of linking phrases with image regions \cite{DBLP:conf/cvpr/XiaoSL17}, but this is not the focus of the present paper.} 

Leveraging images resulted in improved linguistic representations on intrinsic and downstream tasks \cite{multi_sem,autoencoders}.
In most of these approaches, cross-modal projections are learned to incorporate visual semantics in the final representations \cite{Multimodal_Skipgram,guillem, DBLP:conf/naacl/KielaCJN18}.
These works rely on paired textual and visual data and the hypothesis of a one-to-one correspondence between modalities is implicitly assumed: an image of an object univocally represents a word.
However, there is no obvious reason implying that the structure of the two spaces should match. Indeed,  \citet{DBLP:conf/acl/CollellM18} empirically show that cross-modal projection of a source modality does not resemble the target modality in terms of its neighborhood structure. This is  especially  the case for sentences, where many different sentences can describe a similar image. 
Therefore, we argue that learning grounded representations with projections to a visual space is particularly inadequate in the case of sentences.


To overcome this issue, we propose an alternative approach where the structure of the visual space is \textit{partially transferred} to the textual space. 
This is done by distinguishing two types of complementary information sources.
First, the \emph{cluster information}: the implicit knowledge that sentences associated with the same image refer to the same underlying reality.
Second, the \emph{perceptual information}, which is contained within high-level representations of images.
These two sources of information aim at transferring the structure of the visual space to the textual space. Besides, to preserve textual semantics and to avoid an over-constrained textual space, 
 we propose to incorporate the visual information to textual representations using an intermediate representation space that we call \textit{grounded space}, on which cluster and perceptual objectives are trained. 

Our contributions are the following: 
(1) we define two complementary objectives to ground the textual space, based on implicit and explicit visual information;
(2) we propose to incorporate visual semantics through the means of an intermediate space, within which the objectives are learned. 
Moreover (3) we perform quantitative and qualitative evaluations on several transfer tasks, showing the advantages of our approach with respect to previous grounding methods. 

\section{Related work}

Over the last years, several approaches have been proposed to learn semantic representations for sentences.
This includes supervised and task-specific techniques with recursive networks \cite{DBLP:conf/emnlp/SocherPWCMNP13}, convolutional networks \cite{DBLP:conf/acl/KalchbrennerGB14} or self-attentive networks \cite{Attentive,DBLP:conf/emnlp/ConneauKSBB17}, but also unsupervised methods producing universal representations given large text corpora. Examples of the latter include models such as FastSent \cite{fastsent}, QuickThought \cite{quickthought}, Word Information Series \cite{DBLP:journals/csl/Arroyo-Fernandez19}, Universal Sentence Encoder \cite{DBLP:conf/emnlp/CerYKHLJCGYTSK18}, or SkipThought \cite{skip_thought}, where a sentence is encoded with a Gated Recurrent Unit (GRU), and two GRU decoders are trained to reconstruct the adjacent sentences in a dataset of ordered sentences.


To model the way language conveys meaning, traditional approaches consider language as a purely symbolic system based on words and syntactic rules \cite{chomsky, burgess}.
However, \citet{barsalou,Fincher-Kiefer2001} insist on the intuition that language has to be grounded in physical reality and perceptual experience.
The importance of language grounding is underlined in \citet{Gordon:2013:RBK:2509558.2509563}, where an important bias is reported: the frequency at which objects, relations, or events occur in natural language is significantly different from their real-world frequency (e.g., in texts, people are \textit{murdered} four times more than they \textit{breathe}). 
Thus, leveraging visual resources, in addition to textual resources, is a promising way to acquire commonsense knowledge \cite{DBLP:conf/cvpr/LinP15,DBLP:conf/naacl/YatskarOF16} and to cope with the bias between text and reality. 

This intuition has motivated several works for learning visually grounded representations for words, using images --- or abstract scenes \cite{DBLP:conf/cvpr/KotturVMP16}.
Two lines of work can be distinguished.
First, \textit{sequential} techniques combine textual and visual representations that were learned separately \cite{multi_sem,autoencoders,guillem}.
Second, \textit{joint} methods learn a multimodal representation from multiple sources simultaneously. The advantage is that visual information associated with concrete words can be transferred to more abstract ones, which usually have no associated visual data \cite{weighted-gram, Multimodal_Skipgram}.
Closer to our contribution, some approaches learn grounded word embeddings by building upon the skip-gram objective \cite{mikolov} and enforcing word vectors to be close to their corresponding visual features \cite{Multimodal_Skipgram} or their visual context \cite{zablocki} in a multimodal representation space. 
These approaches learn word representations while we specifically target sentences. This task is more challenging since sentences are inherently different than words due to their sequential and compositional nature. Moreover, a great number of different sentences can be generated for the same image. 

\begin{figure*}[t]
    \centering
    \def\svgwidth{\textwidth} 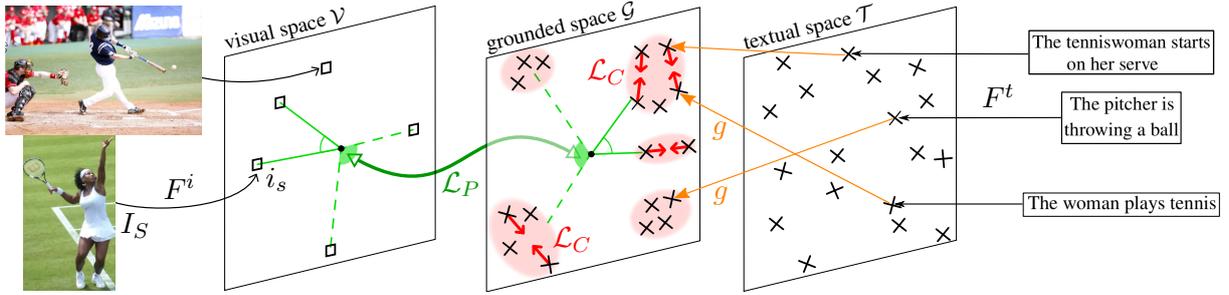
    \caption{\small \label{textual_visual_spaces}
    Model overview.
    Red circles indicate visual clusters. Red arrows represent the gradient of the cluster loss, which gathers visually equivalent sentences --- the contrastive term in loss $\mathcal{L_C}$ is not represented. The green arrow and angles illustrate the perceptual loss, ensuring that cosine similarities correlate across modalities. The origin is at the center of each space.}
    \vspace{-0.4cm}
    \label{fig:overview}
\end{figure*}

Finally, some works learn sentence representations by aligning visual data to sentences within captioning datasets. 
\citet{DBLP:conf/acl/ChrupalaKA15} propose the \texttt{IMAGINET} model where two sentence encoders share word embeddings: a first GRU encoder learns a language model objective and the other one is trained to predict the visual features associated to a sentence. 
The model of \citet{DBLP:conf/naacl/KielaCJN18} is close to \texttt{IMAGINET} and additionally hypothesizes that associated captions ground the meaning of a sentence. 
Both of these works learn a projection of the sentence representation to the corresponding image, which we argue is problematic as it over-constrains the textual space and can degrade textual representations.
Indeed, \citet{DBLP:conf/acl/CollellM18} empirically demonstrated that when a cross-modal mapping is learned, the projection of the source modality does not resemble the target modality, in the sense of nearest neighbor comparison. 
This suggests that cross-modal projections are not appropriate to incorporate visual semantics in text representations. 


\section{Incorporating visual semantics within an intermediate grounded space}

\label{section_model}

\subsection{Model overview}
\label{sec:model_overview}

In this work, we aim at learning grounded representations by jointly leveraging the textual and visual contexts of a sentence.
We note $S$ a sentence and ${s = F^t (S ; \theta^t)}$ its representation computed with a sentence encoder $F^{t}$ parametrized by $\theta^t$.
We follow the classical approach developed in the language grounding literature at the word level \cite{Multimodal_Skipgram,zablocki}, which balances a textual objective $\mathcal{L_T}$ with an additional grounding objective $\mathcal{L_G}$:

\begin{equation} 
\label{eq:global}
\mathcal{L}(\theta^t, \theta^i) = \mathcal{L_{T}}(\theta^t) + \mathcal{L_G}(\theta^t, \theta^i)
\end{equation}

The parameters $\theta^t$ of the sentence encoder $F^t$ are shared in $\mathcal{L_T}$ and $\mathcal{L_G}$, and therefore benefit from both textual and grounding objectives. 
$\theta^i$ denotes extra grounding parameters, including the weights of the image encoder $F^i$. 
Note that any textual objective $\mathcal{L_T}$ and sentence encoder $F^t$ can be used. In our experiments, we choose the well-known SkipThought model \cite{skip_thought}, trained on a corpus of ordered sentences. 

In what follows, we focus on the modeling of the grounding objective $\mathcal{L_G}$, learned on a captioning corpus, where each image is associated with several captions.
Grounding approaches generally leverage visual information by embedding textual and visual elements within the same multimodal space \cite{autoencoders, DBLP:conf/naacl/KielaCJN18}. However, it is not satisfying since texts and images are forced to be in one-to-one correspondence.
Moreover, a caption can
(1) have a wide variety of paraphrases and related sentences describing the same scene (e.g., \textit{the kitten is devouring a mouse} versus \textit{a cat eating a mouse}),
(2) be visually ambiguous (e.g., \textit{a cat is eating} can be associated with many different images, depending on the visual scene/context), or
(3) carry non-visual information (e.g., \textit{cats often think about their meals}). 
Usual grounding objectives, that embed sentences in the visual space, can discard non-visual information (3) through the projection function. 
They can handle (1) by projecting related sentences to the same location in the visual space. However, they are over-sensitive to visual ambiguity (2), because ambiguous sentences should be projected to different locations of the visual space, which is not possible with current grounding models.


To overcome this lack of flexibility, we propose the following approach, illustrated in \autoref{fig:overview}.
To cope with (1), sentences associated with the same image should be close --- we call this \textit{cluster information}. 
To cope with (2), we want to avoid projecting sentences to a particular point of the visual space: instead, we require that the similarity between two images in the visual space (which is linked to the “context discrepancy”) should be close to the similarity between their associated sentences in the textual space.  We call this \textit{perceptual information}.
Finally, as we want to preserve non-visual information in sentence representations (3), we make use of an intermediate space, called \textit{grounded space}, that allows textual representations to benefit from visual properties without degrading the semantics brought by the textual objective $\mathcal{L_T}$.


\subsection{Grounding space and objectives}
\label{visual_context}

In this section, we introduce more formally the grounded space and the different information (cluster and perceptual) captured in the  grounding loss $\mathcal{L_G}$. 

\paragraph{Grounded space}
The grounded space relaxes the assumption that textual and visual representations should be guided by one-to-one correspondences. It rather assumes that the structure of the textual space might be partially modeled on the structure of the visual space.
Thus, instead of directly applying the grounding objectives on a sentence $s$ embedding, we  propose to train the grounding objective $\mathcal{L_G}$ on an intermediate space called \textit{grounded space}. Practically, we use a projection $g(s; \theta^i_g)$ of a sentence $s$ from the textual space to the grounded space. We denote it $g(s)$ for simplicity, where $g$ is a multi-layer perceptron with input $s = F^t(S; \theta^t)$ and parameters $\theta^i_g$  ($\theta^i_g \subset \theta^i$).

\paragraph{Cluster information ($\textbf{C}_g$)} 
The cluster information leverages the fact that two sentences describe, or not, the same underlying reality. In other words, the goal is to  measure if two sentences are \textit{visually equivalent} (assumption (1) in Section 3.1) without considering the content of related images. 
For convenience, two sentences are said to be \textit{visually equivalent} (resp. \textit{visually different}) if they are associated with the same image (resp. different images), i.e.\ if they describe the same (resp. different) underlying reality. We call \textit{cluster} a set of visually equivalent sentences.
For instance, in \autoref{textual_visual_spaces}, 
sentences \textit{The tenniswoman starts on her serve} and \textit{The woman plays tennis} are visually equivalent and belong to the same cluster.

Our hypothesis is that \textit{the similarity between visually equivalent sentences $(s, s^{+})$ should be higher than visually different sentences $(s, s^{-})$.} 
%
We translate this hypothesis into the constraint in the grounded space: ${\cos(g(s),g(s^{+})) \leq  \cos(g(s),g(s^{-}))}$. 
Following \cite{DBLP:conf/cvpr/KarpathyL15, carvalho}, we use a max-margin ranking loss to ensure the gap between both terms is higher than a fixed margin~$\gamma$ (cf.\ red elements in \autoref{textual_visual_spaces}) resulting in the cluster loss $\mathcal{L_{C}}$:
\begin{equation}
\Resize{\columnwidth}{
\mathcal{L}_{C} \hspace{-2pt} = \hspace{-12pt} \sum \limits_{(s,s^{+},s^{-})} \hspace{-4pt} \big\lfloor \hspace{-1pt} \gamma \hspace{-1.5pt} - \hspace{-1pt} \cos(g(s),g(s^{\hspace{-1pt}+})\hspace{-2pt}) + \cos(g(s),g(s^{\hspace{-1pt}-}\hspace{-1pt})\hspace{-2pt}) \hspace{-2pt} \big\rfloor_{\hspace{-2pt}+}
}
\end{equation}
where $s^{+}$ (resp. $s^{-}$) is a randomly sampled visually equivalent (resp. different) sentence to $s$. This loss function is also used in the cross-modal retrieval literature to enforce structure-preserving constraints between sentences describing a same image \cite{DBLP:conf/cvpr/WangLL16}. 

\paragraph{Perceptual information ($\textbf{P}_g$)} The cluster hypothesis alone ignores the structure of the visual space and only uses the visual modality as a proxy to assess if two sentences are visually equivalent or different.  Moreover, the ranking loss $\mathcal{L_C}$ simply drives apart visually different sentences in the representation space, which can be a problem when two images have a closely related content. For instance, the baseball and tennis images in \autoref{textual_visual_spaces} may be different, but they are both sports images, and thus their corresponding sentences should be somehow close in the grounded space. Finally, it supposes that we have a dataset of images associated with several captions.

To cope with these limitations, we consider the structure of the visual space and use the content of images. 
The intuition is that the structure of the textual space should be modeled on the structure of the visual one to extract visual semantics.
We choose to preserve \textit{similarities} between related elements across spaces (cf.\ green elements in \autoref{textual_visual_spaces}). We  thus assume that
\textit{the similarity between two sentences in the grounded space should be correlated with the similarity between their corresponding images in the visual space.} 
We translate this hypothesis into the perceptual loss $\mathcal{L_{P}}$: 
\begin{equation}
    \mathcal{L_P}=-\rho(\{\textit{sim}^\text{text}_{k_1,k_2}\}, \{\textit{sim}^\text{im}_{k_1,k_2}\})
\end{equation}
where $\rho$ is the Pearson correlation, 
$\textit{sim}^\text{text}_{k_1,k_2}=\cos(g(s_{k_1}), g(s_{k_2}))$
and 
$\textit{sim}^\text{im}_{k_1,k_2}=\cos(i_{k_1}, i_{k_2})$
are respectively textual and visual similarities computed over several randomly sampled pairs of matching sentences and images.




\paragraph{Grounded loss} 
Taking altogether, the grounded space and cluster/perceptual information leads to the grounding objective $\mathcal{L_{G}}(\theta^t, \theta^i)$ as a linear combination of the aforementioned objectives:
\begin{equation}
\mathcal{L_{G}}(\theta^t, \theta^i) = 
\alpha_C \mathcal{L}_C(\theta^t, \theta^i)
+ 
\alpha_P \mathcal{L}_P(\theta^t, \theta^i) 
\end{equation}
where $\alpha_C$ and $\alpha_P$ are hyper-parameters weighting contributions of $\mathcal{L}_C$ and $\mathcal{L}_P$. $\theta^i$ corresponds to all the grounding-related parameters, i.e. those of the image encoder $F^i$ and of the projection function $g$ (i.e., $\theta^i_g$).

\section{Evaluation protocol}
\subsection{Datasets}
\label{datasets}
\paragraph{Textual dataset.} Following~\cite{skip_thought,fastsent}, we use the Toronto BookCorpus dataset as the textual corpus. This corpus consists of 11K books, and 74M ordered sentences, with an average of 13 words per sentence. 

\paragraph{Visual dataset.} We use the MS COCO \cite{mscoco} dataset as the visual corpus. This image captioning dataset consists of 118K/5K/41K (train/val/test) images, each with five English descriptions. Note that the number of sentences in the training set of COCO (590K sentences) only represents $0.8 \%$ of the sentence data in BookCorpus, which is negligible, and the additional textual training data cannot account for performance discrepancies between textual and grounded models.


\subsection{Baselines and Scenarios}
\label{baselines}

In the experiments, we focus on one of the most established sentence models: SkipThought (noted $\textbf{T}$) as the textual baseline: the parameters of the sentence embedding model are obtained by minimizing $\mathcal{L_T}$. 
Then, we derive several baselines and scenarios based on \textbf{T}, each representing a different approach of grounding. Since our focus is to study the impact of grounding on sentence representations, all baselines and scenarios share the same representation dimension $d_t=2048$ and are trained on the same datasets (cf.\ \autoref{datasets}). We also report a textual model of dimension $\frac{d_t}{2}$ that we call $\textbf{T}_{1024}$, to compare with the GroundSent model of \cite{DBLP:conf/naacl/KielaCJN18}. 

\paragraph{Model Scenarios.}
We test variants of our grounding model presented in \autoref{section_model}, all based on $\textbf{T}$:  $\textbf{T}+\textbf{C}_g$, $\textbf{T}+\textbf{P}_g$, $\textbf{T}+\textbf{C}_g+\textbf{P}_g$, where $\textbf{C}_g$ (resp. $\textbf{P}_g$) represents the loss $\mathcal{L}_C$ (resp. $\mathcal{L}_P$). We also consider scenarios where $g$ equals the identity function (no grounded space), which we note $\textbf{C}_{id}$, $\textbf{P}_{id}$, $\textbf{C}_{id}+\textbf{P}_{id}$, etc. 
Finally, we also performed preliminary analysis learning only from the visual modality: $\textbf{C}_{g/id}$, $\textbf{P}_{g/id}$, $\textbf{C}_{g/id}+\textbf{P}_{g/id}$. 
 
 
\paragraph{Baselines.}
We adapt two classical multimodal word embedding models for sentences. Accordingly, models from the two existing model families are considered:
\newlength{\ltwosmall}
\setlength{\ltwosmall}{25pt}

\begin{table*}[t]
\small
\begin{center}

\begin{tabularx}{\textwidth}{l | Y Y Y Y | Y Y Y Y Y} \multicolumn{1}{r}{} &  \multicolumn{4}{c|}{\textbf{Structural measures}} & \multicolumn{5}{c}{\textbf{Semantic relatedness}} \\
\multicolumn{1}{c|}{Model} & $\mnno$ & $\rho_\textit{vis}$& $C_\textit{inter}$ & $C_\textit{intra}$ & STS/All & STS/Cap & STS/News &STS/Forum & SICK  \\
\hline
$\textbf{T}$ &10.0 & 4.1 & 54.2 & 70.1& 30 &41&36 &21 &51\\
\textbf{CM} (text)& 24.2&12.8&41.7 &74.8& 52&76&42&\textbf{37}&55 \\ 
$\textbf{P}_{id}$& 21.1&\textbf{37.9}&42.2&69.3&45 &66&41&34&54\\ 
$\textbf{C}_{id}$&27.5&10.5&\textbf{2.9}&\textbf{84.7} &60&83&45&20&55\\ 
$\textbf{C}_{id}+\textbf{P}_{id}$&\textbf{27.9}&25.8&6.7&82.6& \textbf{61}&\textbf{84}&\textbf{46}&28&\textbf{57}\\ 

\hline 
\textbf{CM} (vis.)& 27.1 & 19.2 & 1.5 & 85.8 & 56 & 78 & 40 & 34 & 55 \\ 
$\textbf{P}_g$& 21.3& \textbf{32.4}& 43.9&73.3& 45&66&41&\textbf{37}&53\\ 
$\textbf{C}_g$& 28.6 & 9.4&\textbf{1.1}& \textbf{88.5}& 62&83&46&29&59\\ 
$\textbf{C}_g+\textbf{P}_g$& \textbf{28.9} & 29.1& 4.7&87.5& \textbf{63} &\textbf{84}& \textbf{48}&33&\textbf{60}\\

\end{tabularx}
\end{center}
\vspace{-0.28cm}

\caption{Intrinsic evaluations carried out on the grounded space for models with $g=\text{MLP}$; the textual space for \textbf{T}, \textbf{CM} (text) and models with $g=id$; and the visual space for \textbf{CM} (vis).}
\vspace{-0.48cm}
\label{vrep_M2}
\end{table*}

\noindent\emph{Cross-modal Projection} (\textbf{CM}): Inspired by \citet{Multimodal_Skipgram}, this baseline learns to project sentences in the visual space using a max-margin loss:

\vspace{-0.1cm}
\begin{equation*}
\sum \limits_{(s, i_s, i^-)} \big\lfloor \gamma' + \cos(f(s),i^{-}) - \cos(f(s),i_{s}) \big\rfloor_+
\end{equation*}
 \vspace{-0.1cm}

\noindent where $f$ is a MLP, 
$\gamma'$ a fixed margin and $i^-$ a non-matching image. Similarly to our scenarios, the sentence encoder is initialized with \textbf{T}. \\ 

\noindent \emph{Sequential} (\textbf{SEQ}): Inspired by \citet{guillem}, we learn a linear regression model $(W,b)$ to predict the visual representation of an image, from the representation of a matching caption.
The grounded word embedding is the concatenation of the original SkipThought vector \textbf{T} and its predicted (``\textit{imagined}'') representation $W\textbf{T}+b$, which is projected using a PCA into dimension $d_t$. 

In both cases, the parameters to be learned, in addition to the sentence encoder, are the cross-modal projections -- and the sentence representation is obtained by averaging word vectors.

\paragraph{GroundSent Model} We re-implement the GroundSent models of \citet{DBLP:conf/naacl/KielaCJN18}, obtaining comparable results.
The authors propose two objectives to learn a grounded vector: (a) Cap2Img: the cross-modal projections of sentences are pushed towards their respective images via a max-margin ranking loss, and (b) Cap2Cap: a visually equivalent sentence is predicted via a LSTM sentence decoder. The Cap2Both objective is a combination of these two objectives. Once the grounded vectors are learned, they are concatenated with a textual vector (learned via a SkipThought objective) to form the GS-Img, GS-Cap and GS-Both vectors. 


\vspace{-0.1cm}
\subsection{Evaluation tasks and metrics}
\label{tasks}

In line with previous works \cite{skip_thought,fastsent}, we consider several benchmarks to evaluate the quality of our grounded embeddings:

\paragraph{Semantic relatedness.} We use two semantic similarity benchmarks: STS \cite{DBLP:conf/semeval/CerDALS17} and SICK \cite{SICK}, which consist of pairs of sentences that are associated with human-labeled similarity scores. STS is subdivided into three textual sources: \textit{Captions} contain concrete sentences describing daily-life actions, whereas the others contain more abstract sentences: news headlines in \textit{News} and posts from user forums in \textit{Forum}.
The Spearman correlations are measured between the cosine similarity of our learned sentence embeddings and human-labeled scores. 

\paragraph{Classification benchmarks.} All extrinsic evaluations are carried out using the SentEval pipeline \cite{conneau2018senteval}. The tasks are the following: opinion polarity (MPQA) ~\cite{MPQA}, movie review sentiment (MR) ~\cite{MR}, subjectivity/objectivity classification (SUBJ) ~\cite{SUBJ}, customer reviews (CR) \cite{CR}, binary sentiment analysis on SST \cite{DBLP:conf/emnlp/SocherPWCMNP13}, paraphrase identification (MSRP) \cite{MSRP} as well as two entailment classification benchmarks: SNLI \cite{DBLP:conf/emnlp/BowmanAPM15} 
and SICK \cite{DBLP:conf/lrec/MarelliMBBBZ14}. For each dataset, a logistic regression classifier is learned from the extracted sentence embeddings, and we report the classification accuracy.


\vspace{-0.1cm}
\paragraph{Structural measures.} To probe the learned grounded space, we define structural measures, and report their values on the validation set of MS COCO (5K images, 25K captions). First, we report the \textit{mean Nearest Neighbor Overlap} ($\mnno$) metric, as defined in \citet{DBLP:conf/acl/CollellM18}, that indicates the proportion of shared nearest neighbors between image representations and their corresponding captions in their respective spaces. To study \emph{perceptual information}, we define $\rho_\textit{vis}$, the Pearson correlation $\rho(\cos(s,s'), \cos(v_{s},v_{s'}))$ between images and their corresponding sentences' similarities. For \emph{cluster information}, we introduce $C_\textit{intra} = \mathbb{E}_{v_{s}=v_{s'}}[cos(s,s')] $, which measures the homogeneity of each cluster,
and  $C_\textit{inter} = \mathbb{E}_{v_{s} \neq v_{s'}}[cos(s,s')] $, which measures how well clusters are separated from each other.

\vspace{-0.1cm}
\subsection{Implementation details}
\label{sec:implementation-details}

Images are processed using a pretrained Inception-v3 network \cite{DBLP:conf/cvpr/SzegedyVISW16} ($d_i=2048$).
The model is trained with ADAM \cite{DBLP:journals/corr/KingmaB14} and a learning rate $l_r=8.10^{-4}$. 
As done in \citet{skip_thought}, our sentence encoder is a GRU with a vocabulary of $20$K words, represented in dimension 620; we perform vocabulary expansion at inference. 
All hyperparameters are tuned using the Pearson correlation measure on the validation set of the SICK benchmark:  $\gamma=\gamma'=0.5$, $\alpha_C=\alpha_P=0.01$, $d_g=512$; functions $f$ and $g$ are 2-layer MLP. 
As done in \cite{DBLP:conf/naacl/KielaCJN18}, we set $d_t= 2048$. 

\section{Experiments and Results}

Our main objective is to study the contribution brought by the visual modality to the grounded sentence representations, and we do not attempt to outperform purely textual sentence encoders from the literature.
We show that textual models can benefit from grounding approaches without requiring any changes to the original textual objectives $\mathcal{L}_T$. We report quantitative and qualitative insights 
(\autoref{sec:results:insights}), and quantitative results 
on the SentEval benchmark 
(\autoref{sec:results:quantitative}). 


\begin{figure*}[t]
\centering
   \includegraphics[width=\linewidth]{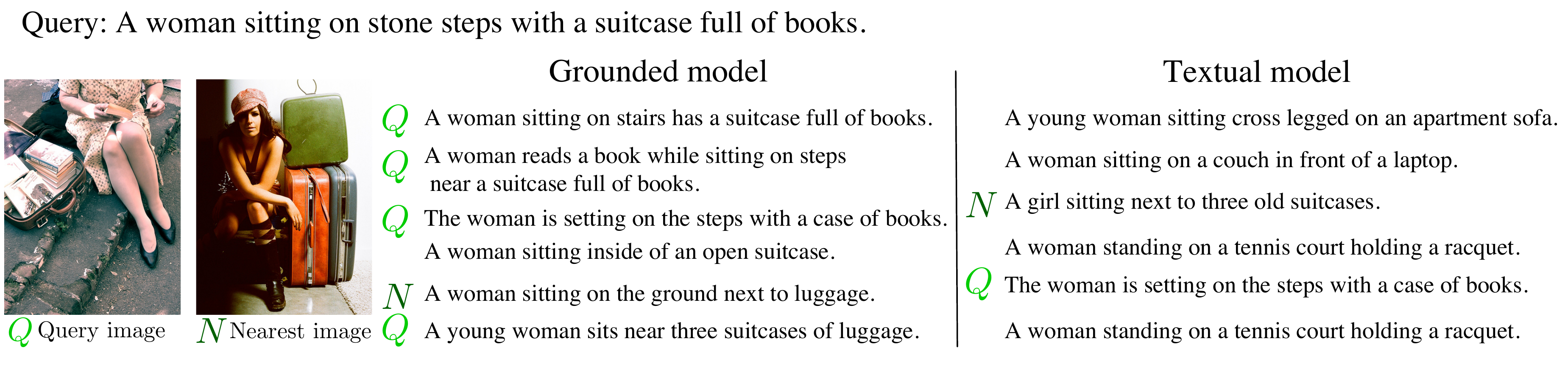}
       \vspace{-0.9cm}
    \caption{Nearest neighbors of a selected sentence in the validation set of MS COCO, for both grounded and purely textual models. $Q$ is the query image, $N$ is the nearest neighbor of $Q$ in the visual space. Sentences that are caption of $Q$ or $N$ are prefixed with $Q$ or $N$.}
    \label{mnno_fig}
    \vspace{-0.1cm}
\end{figure*} 

\begin{figure}[t]
\centering
   \includegraphics[width=\columnwidth]{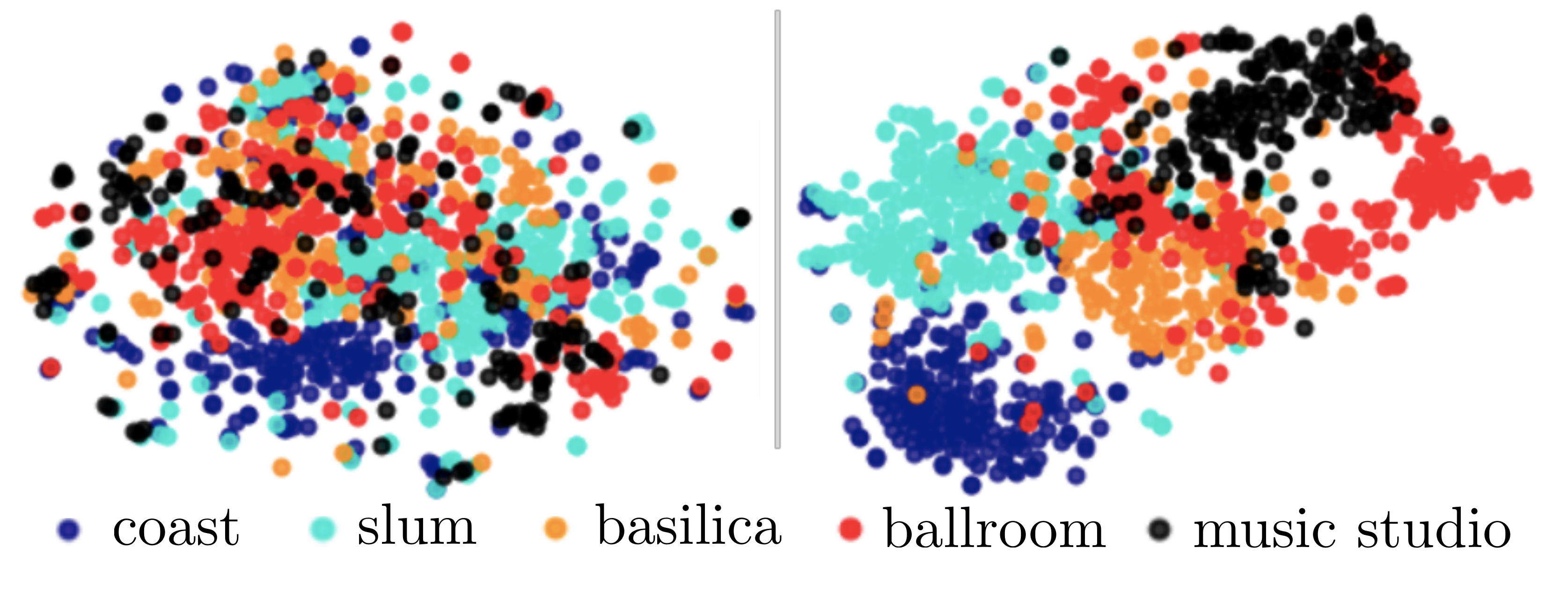}
   \vspace{-1.0cm}
    \caption{t-SNE visualization on CMPlaces sentences for a set of randomly sampled visual scenes. Left: textual model \textbf{T}. Right: grounded model $\textbf{C}_g+\textbf{P}_g$.}
    \label{tsne}
    \vspace{-0.6cm}
\end{figure}

\newlength{\fsize}
\setlength{\fsize}{9.2pt}
\newlength{\flsize}
\setlength{\flsize}{10pt}
\begin{table*}[t]
    \begin{center}
    \begin{tabularx}{\textwidth}{@{} l | c c}
      \multicolumn{1}{c|}{Query} & \multicolumn{1}{c}{Textual model} & \multicolumn{1}{c}{Grounded model} \\
      \hline 
{\fontsize{\fsize}{\flsize}\selectfont Two people are in \textbf{love}} & {\fontsize{\fsize}{\flsize}\selectfont Two people are fencing indoors} & {\fontsize{\fsize}{\flsize}\selectfont A couple just got married and are taking a picture with family} \\
{\fontsize{\fsize}{\flsize}\selectfont A man is \textbf{horrified}} & {\fontsize{\fsize}{\flsize}\selectfont A man and a woman are smiling} & {\fontsize{\fsize}{\flsize}\selectfont A teenage boy wearing a cap looks irritated} \\
{\fontsize{\fsize}{\flsize}\selectfont This is a \textbf{tragedy}} & {\fontsize{\fsize}{\flsize}\selectfont A group of people are at a party} & {\fontsize{\fsize}{\flsize}\selectfont Men doing a war reenactment} \\

    \end{tabularx}
    \vspace{-0.2cm}
    \caption{Qualitative analysis: nearest neighbor of a given query (containing an abstract word) among Flickr30K sentences.}
    \label{table_qualitative}
    \end{center}
   \vspace{-0.6cm}
\end{table*}

\newlength{\lsmallbig}
\setlength{\lsmallbig}{27pt}
\newlength{\lverysmall}
\setlength{\lverysmall}{7.7pt}
\newlength{\lcolumn}
\setlength{\lcolumn}{40pt}

\begin{table*}[t]
    \begin{center}
    \footnotesize
	\begin{tabularx}{\textwidth}{l l | Y Y Y Y C{\lcolumn} Y Y Y  Y}
    	\multicolumn{2}{c|}{Model} & {\fontsize{\lverysmall}{0}\selectfont MR} & {\fontsize{\lverysmall}{0}\selectfont CR} & {\fontsize{\lverysmall}{0}\selectfont SUBJ} & {\fontsize{\lverysmall}{0}\selectfont MPQA} &  {\fontsize{\lverysmall}{0}\selectfont MRPC} & {\fontsize{\lverysmall}{0}\selectfont SST} & {\fontsize{\lverysmall}{0}\selectfont SNLI} & {\fontsize{\lverysmall}{0}\selectfont SICK} & {\fontsize{\lverysmall}{0}\selectfont AVG}	\\
    	\hline

\cite{skip_thought}$^\dag$ & $\textbf{T}_{1024}$&72.7$^{*}$ & 75.2$^{*}$ & 90.6$^{*}$ & 84.7$^{*}$ & 71.8$^{*}$/79.2$^{*}$ & 76.2$^{*}$ & 68.8$^{*}$ & 79.3$^{*}$& 77.4 \\

\cite{DBLP:conf/naacl/KielaCJN18}$^\dag$ & $\text{GS-Cap}$& 72.0$^{*}$ & 76.8$^{*}$ & 90.7$^{*}$ & 85.5$^{*}$ & 72.9/80.6 & 76.7$^{*}$ & 73.7 & \textbf{82.9} & 78.4 \\
\cite{DBLP:conf/naacl/KielaCJN18}$^\dag$ &$\text{GS-Img}$& 74.5$^{*}$ & 79.3$^{*}$ & 90.8$^{*}$ & 87.8$^{*}$ & 73.0/80.3 & 80.0$^{*}$ & 72.2$^{*}$ & 80.9$^{*}$ & 79.8\\

\cite{DBLP:conf/naacl/KielaCJN18}$^\dag$ & $\text{GS-Both}$ & 72.5$^{*}$ & 75.7$^{*}$ & 90.7$^{*}$ & 85.4$^{*}$ & 72.9/81.3 & 76.7$^{*}$ & 72.2$^{*}$ & 81.4$^{*}$ & 78.4\\
\hline
\cite{skip_thought}$^\dag$ & $\textbf{T}$& 75.9$^{*}$& 79.2$^{*}$ & 92.0 & 86.7$^{*}$ & 72.2/80.2 & 81.8$^{*}$ & 72.0$^{*}$ & 81.1$^{*}$ & 80.1 \\

\cite{lazaridou}$^\ddag$ & $\textbf{T}+\textbf{CM}$& 77.6 & 81.4 & 92.6 & 88.3 & 73.5/81.1 & 82.0$^{*}$ & 73.0 & 81.4$^{*}$ & 81.1 \\
\cite{guillem}$^\ddag$ & $\textbf{SEQ}$ & 76.1$^{*}$& 79.8$^{*}$&92.5&86.7$^{*}$&70.0$^{*}$/79.5$^{*}$&81.7$^{*}$&  67.3$^{*}$& 76.7$^{*}$ & 78.9 \\

\hline

\multirow{6}{*}{Model scenarios} & $\textbf{T}+\textbf{P}_{id}$& 77.5 & 81.5 & 92.7 & 88.4 & \textbf{73.7}/81.3 & 82.4 & 72.4 & 81.1 & 81.2\\
& $\textbf{T}+\textbf{P}_g$&\textbf{77.8} & \textbf{81.8} & \textbf{93.0} & 88.1 & 73.3/\textbf{81.6} & \textbf{83.5} & 72.8 & 82.2 & \textbf{81.6} \\ 
& $\textbf{T}+\textbf{C}_{id}$ & 77.5 & 81.6 & 92.8 & 88.3 & 72.9/80.5 & 82.2 & 73.1 & 82.3 & 81.3 \\
&$\textbf{T}+\textbf{C}_g$& 77.3 & 81.5 & 92.8 & \textbf{88.6} & 73.6/81.1 & 82.6 & \textbf{74.1} & 82.6 & \textbf{81.6} \\ 
& $\textbf{T}+\textbf{C}_{id}+\textbf{P}_{id}$ & 77.3 & 81.2& \textbf{93.0} & 88.4 & 73.0/80.6 & 82.5&73.5 &82.1 & 81.4 \\ 
&$\textbf{T}+\textbf{C}_g + \textbf{P}_g$& 77.4 & 81.5 & \textbf{93.0} & 88.1 & 73.2/80.9 & 82.7 & 73.9 & \textbf{82.9} & \textbf{81.6} \\ 

    \end{tabularx}
    \vspace{-0.3cm}
    \caption{Extrinsic evaluations with SentEval. {\small All models give sentences in dimension $d_t=2048$ (except $\mathbf{T}_{1024}$). `AVG' stands for the average accuracies reported in the other columns.
    `$\dag$': the model has been re-implemented (we obtained higher scores than the one given in the original papers). `$\ddag$': the baseline is an adaptation of the model to the case of sentences. '$*$': significantly differs from the best scenario among our models.}
    \label{table:senteval}}
    \vspace{-0.7cm}
    \end{center}
\end{table*}

\subsection{Study of the grounded space}
\label{sec:results:insights}

We study the impact of the various grounding hypotheses on the structure of the grounded space, using intrinsic measures. 
In \autoref{vrep_M2}, we report the structural measures and the semantic relatedness scores of the baselines, namely \textbf{T} and \textbf{CM}, and on 
the various scenarios of our model. The textual loss is discarded to isolate the effect of the different grounding hypotheses.




\paragraph{The impact of grounding}
We investigate the effect of grounding on sentence representations. 
Results highlight that all grounded models improve over the baseline \textbf{T}. Moreover, our model $\textbf{C}_g+\textbf{P}_g$ is generally the most effective regarding the $\mnno$ measure and semantic relatedness tasks. 

\textit{Influence of concreteness \hspace{0.4cm}} To understand in which cases grounding is useful, we compute the average visual concreteness $\bar{c}$ of the STS benchmark, which is divided in three categories (\textit{Captions}, \textit{News}, \textit{Forum}). This is done by using a concreteness dataset built by \citet{concreteness} consisting of human ratings of concreteness (between $0$ and $5$) for 40,000 English words; for a given benchmark, we compute the sum of these scores and average over all words that are in the concreteness dataset. 
The performance gain $\Delta$ between $\textbf{C}_g+\textbf{P}_g$ and  \textbf{T} are observed when the visual concreteness $\bar{c}$ is high: for \textit{Captions} ($\bar{c}=3.10$), the improvement is substantial: ($\Delta=+43$); for benchmarks with a lower concreteness (\textit{News} with $\bar{c}=2.61$ and \textit{Forum}  with $\bar{c}=2.39$), the improvement is smaller ($\Delta=+12$). Thus, grounding brings useful complementary information, especially for concrete sentences. 

\textit{t-SNE visualization \hspace{0.5cm}} 
This finding is also supported by a qualitative experiment showing that grounding groups together similar visual situations.
Using sentences from CMPlaces \cite{castrejon2016learning}, which describe visual scenes (e.g., \textit{coast}, \textit{shoe-shop}, \textit{plaza}, etc.) and are classified in 205 scene categories, we randomly sample 5 visual scenes and plot in \autoref{tsne} the corresponding sentences using t-SNE \cite{maaten2008visualizing}. We notice that our grounded model is better able to cluster sentences that have a close visual meaning than the text-only model. This is reinforced by the structural measures computed on the five clusters of \autoref{tsne}: $C_\textit{inter}=19,C_\textit{intra}=22$ for \textbf{T}, $C_\textit{inter}=11,C_\textit{intra}=27$ for $\textbf{C}_g+\textbf{P}_g$. Indeed, $C_\textit{inter}$ (resp. $C_\textit{intra}$), is lower (resp. higher) for the grounded model $\textbf{C}_g+\textbf{P}_g$ compared to  \textbf{T}, which shows that clusters corresponding to different scenes are more clearly separated (resp. sentences corresponding to a given scene are more packed).

\textit{Nearest neighbors search \hspace{0.5cm}} 
Furthermore, we show in \autoref{table_qualitative} that concrete knowledge acquired via our grounded model can also be transferred to abstract sentences. To do so, we manually build sentences using words with low concreteness (between 2.5 and 3.5) from the USF dataset \cite{usf}. Then, nearest neighbors are retrieved from the set of sentences of Flickr30K \cite{DBLP:conf/iccv/PlummerWCCHL15}. In this sample, we see that our grounded model is more accurate than the purely textual model to capture visual meaning. 
The observation that visual information propagates from concrete sentences to abstract ones is analogous to findings made in previous research on word embeddings \cite{DBLP:conf/emnlp/HillK14}.

\textit{Neighboring structure \hspace{0.5cm}} 
To illustrate the discrepancy on the $\mnno$ metric observed between $\textbf{C}_g + \textbf{P}_g$ and \textbf{T}, 
we select a query image $Q$ in the validation set of MS COCO, along with its corresponding caption $S$; we display, in \autoref{mnno_fig}, the nearest neighbor of $Q$ in the visual space, noted $N$, and the nearest neighbors of $S$ in the grounded space. With our grounded model, the neighborhood $S$ is mostly made of sentences corresponding to $Q$ or $N$. 


\paragraph{Hypotheses validation} We now validate our hypotheses (cf.\ \autoref{sec:model_overview}) on the grounded space, using the Cross-Modal Projection baseline (\textbf{CM})  and our model scenarios as outlined in \autoref{vrep_M2}.
For fair comparison, metrics for the baseline \textbf{CM} are estimated either  on the visual  or the textual space depending on whether our models rely on the grounded space ($g$) or not ($id$). 
These results correspond to the rows \textbf{CM} (text) and \textbf{CM} (vis.) in \autoref{vrep_M2}. 

Results highlight that:
(1) Using a grounded space is beneficial; indeed, semantic relatedness and $\mnno$ scores are higher in the lower half of \autoref{vrep_M2}, e.g., $\textbf{C}_g > \textbf{C}_{id}$, $\textbf{P}_g > \textbf{P}_{id}$ and $\textbf{C}_g + \textbf{P}_g > \textbf{C}_{id} + \textbf{P}_{id}$; 
(2) Solely using cluster information leads to the highest $C_\textit{intra}$ and lowest $C_\textit{inter}$, which suggests that $\textbf{C}_\bullet$ is the most efficient model at separating visually different sentences;
(3) Using only perceptual information in $\textbf{P}_\bullet$ logically leads to highly correlated textual and visual spaces (highest $\rho_{vis}$), but the local neighborhood structure is not well preserved (lowest $C_\textit{intra}$); 
(4) Our model $\textbf{C}_\bullet+\textbf{P}_\bullet$ is better than $\textbf{CM}$ at capturing cluster information (higher $C_\textit{intra}$, lower $C_\textit{inter}$) and perceptual information (higher $\rho_{vis}$). This also translates in a higher $\mnno$ measure for $\textbf{C}_\bullet+\textbf{P}_\bullet$, leading us to think that the conjunction of both perceptual and cluster information leads to high correlation of modalities, in terms of neighborhood structure. 
Moreover, this high $\mnno$ score results in better performances for our model $\textbf{C}_\bullet+\textbf{P}_\bullet$ in terms of semantic relatedness. 

\subsection{Evaluation on transfer tasks}
\label{sec:results:quantitative}

We now focus on extrinsic evaluation of the embeddings.
\autoref{table:senteval} reports evaluations of our baselines and scenarios on SentEval \citep{conneau2018senteval}, a classical benchmark used for evaluating sentence embeddings. Before further analysis, we find that our grounded models systematically outperform the textual baseline \textbf{T}, on all benchmarks, which shows the first substantial improvement brought by grounding and visual information in a sentence representation model.
Indeed, models GS-Cap, GS-Img and GS-Both from \cite{DBLP:conf/naacl/KielaCJN18}, despite improving over $\textbf{T}_{1024}$, perform worse than the textual model of the same dimension $\textbf{T}$ --- this is consistent with what they report in their paper.

Our results interpretation is the following:
(1) our joint approach shows superior performances over the sequential one, confirming results reported at the word level \cite{zablocki}. Indeed, both sequential models, GS models \cite{DBLP:conf/naacl/KielaCJN18} and \textbf{SEQ} (inspired from \cite{guillem}) are systematically worse than our grounded models for all benchmarks.
(2) Preserving the structure of the visual space is more effective than learning cross-modal projections; indeed, all our models outperform $\textbf{T}+\textbf{CM}$ on average (`AVG' column).
(3) Making use of a grounded space yields slightly improved sentence representations. Indeed, 
our models that use the grounded space ($g=\text{MLP}$) can take advantage of more expression power provided by the trainable $g$ than models which integrate grounded information directly in the textual space ($g=id$). 
(4) 
Among our model scenarios, $\textbf{T}+\textbf{P}_g$ has maximal scores on the most tasks; however, it shows lower scores on SNLI and SICK, which are entailment tasks.
Models using cluster information $\textbf{C}_g$ are naturally more suited for these tasks and hence obtain higher results. Finally, the combined model $\textbf{T}+\textbf{C}_g+\textbf{P}_g$ shows a good balance between classification and entailment tasks. 



\section{Conclusion}
We proposed a multimodal model aiming at preserving the structure of visual and textual spaces to learn grounded sentence representations.
Our contributions include
(1) leveraging both perceptual and cluster information and 
(2) using an intermediate grounded space enabling to relax the constraints on the textual space. 
Our approach is the first to report consistent positive results against purely textual baselines on a variety of natural language tasks.
As future work, we plan to use visual information to specifically target complex downstream tasks  requiring commonsense and reasoning such as question answering or visual dialogue.

\section*{Acknowledgments}
This  work  is  partially  supported  by  the  CHIST-ERA  EU project MUSTER (ANR-15-CHR2-0005) and the Labex SMART  (ANR-11-LABX-65)  supported  by  French  state funds  managed  by  the  ANR  within  the  Investissements d'Avenir program under reference ANR-11-IDEX-0004-02.

\bibliography{biblio}

\begin{thebibliography}{57}
\expandafter\ifx\csname natexlab\endcsname\relax\def\natexlab#1{#1}\fi

\bibitem[{Arroyo{-}Fern{\'{a}}ndez et~al.(2019)Arroyo{-}Fern{\'{a}}ndez,
  M{\'{e}}ndez{-}Cruz, Sierra, Torres{-}Moreno, and
  Sidorov}]{DBLP:journals/csl/Arroyo-Fernandez19}
Ignacio Arroyo{-}Fern{\'{a}}ndez, Carlos{-}Francisco M{\'{e}}ndez{-}Cruz,
  Gerardo Sierra, Juan{-}Manuel Torres{-}Moreno, and Grigori Sidorov. 2019.
\newblock \href {https://doi.org/10.1016/j.csl.2019.01.005} {Unsupervised
  sentence representations as word information series: Revisiting {TF-IDF}}.
\newblock \emph{Computer Speech {\&} Language}, 56:107--129.

\bibitem[{Bahdanau et~al.(2014)Bahdanau, Cho, and
  Bengio}]{DBLP:journals/corr/BahdanauCB14}
Dzmitry Bahdanau, Kyunghyun Cho, and Yoshua Bengio. 2014.
\newblock \href {http://arxiv.org/abs/1409.0473} {Neural machine translation by
  jointly learning to align and translate}.
\newblock \emph{CoRR}, abs/1409.0473.

\bibitem[{Baroni(2016)}]{baroni}
Marco Baroni. 2016.
\newblock \href {https://doi.org/10.1111/lnc3.12170} {Grounding distributional
  semantics in the visual world}.
\newblock \emph{Language and Linguistics Compass}, 10(1):3--13.

\bibitem[{Bowman et~al.(2015)Bowman, Angeli, Potts, and
  Manning}]{DBLP:conf/emnlp/BowmanAPM15}
Samuel~R. Bowman, Gabor Angeli, Christopher Potts, and Christopher~D. Manning.
  2015.
\newblock \href {http://aclweb.org/anthology/D/D15/D15-1075.pdf} {A large
  annotated corpus for learning natural language inference}.
\newblock In \emph{Proceedings of the 2015 Conference on Empirical Methods in
  Natural Language Processing, {EMNLP} 2015, Lisbon, Portugal, September 17-21,
  2015}, pages 632--642.

\bibitem[{Bruni et~al.(2014)Bruni, Tran, and Baroni}]{multi_sem}
Elia Bruni, Nam~Khanh Tran, and Marco Baroni. 2014.
\newblock \href {http://dl.acm.org/citation.cfm?id=2655713.2655714} {Multimodal
  distributional semantics}.
\newblock \emph{J. Artif. Int. Res.}, 49(1):1--47.

\bibitem[{Brysbaert et~al.(2013)Brysbaert, Beth~Warriner, and
  Kuperman}]{concreteness}
Marc Brysbaert, Amy Beth~Warriner, and Victor Kuperman. 2013.
\newblock \href {https://doi.org/10.3758/s13428-013-0403-5} {Concreteness
  ratings for 40 thousand generally known english word lemmas}.
\newblock \emph{Behavior research methods}, 46.

\bibitem[{Burgess and Lund(1997)}]{burgess}
Curt Burgess and Kevin Lund. 1997.
\newblock Modelling parsing constraints with high-dimensional context space.
\newblock 12.

\bibitem[{Carvalho et~al.(2018)Carvalho, Cad{\`{e}}ne, Picard, Soulier, Thome,
  and Cord}]{carvalho}
Micael Carvalho, R{\'{e}}mi Cad{\`{e}}ne, David Picard, Laure Soulier, Nicolas
  Thome, and Matthieu Cord. 2018.
\newblock \href {https://doi.org/10.1145/3209978.3210036} {Cross-modal
  retrieval in the cooking context: Learning semantic text-image embeddings}.
\newblock In \emph{The 41st International {ACM} {SIGIR} Conference on Research
  {\&} Development in Information Retrieval, {SIGIR} 2018, Ann Arbor, MI, USA,
  July 08-12, 2018}, pages 35--44.

\bibitem[{Castrejon et~al.(2016)Castrejon, Aytar, Vondrick, Pirsiavash, and
  Torralba}]{castrejon2016learning}
Lluis Castrejon, Yusuf Aytar, Carl Vondrick, Hamed Pirsiavash, and Antonio
  Torralba. 2016.
\newblock Learning aligned cross-modal representations from weakly aligned
  data.
\newblock In \emph{Computer Vision and Pattern Recognition (CVPR), 2016 IEEE
  Conference on}. IEEE.

\bibitem[{Cer et~al.(2018)Cer, Yang, Kong, Hua, Limtiaco, John, Constant,
  Guajardo{-}Cespedes, Yuan, Tar, Strope, and
  Kurzweil}]{DBLP:conf/emnlp/CerYKHLJCGYTSK18}
Daniel Cer, Yinfei Yang, Sheng{-}yi Kong, Nan Hua, Nicole Limtiaco, Rhomni~St.
  John, Noah Constant, Mario Guajardo{-}Cespedes, Steve Yuan, Chris Tar, Brian
  Strope, and Ray Kurzweil. 2018.
\newblock \href {https://aclanthology.info/papers/D18-2029/d18-2029} {Universal
  sentence encoder for english}.
\newblock In \emph{Proceedings of the 2018 Conference on Empirical Methods in
  Natural Language Processing, {EMNLP} 2018: System Demonstrations, Brussels,
  Belgium, October 31 - November 4, 2018}, pages 169--174.

\bibitem[{Cer et~al.(2017)Cer, Diab, Agirre, Lopez{-}Gazpio, and
  Specia}]{DBLP:conf/semeval/CerDALS17}
Daniel~M. Cer, Mona~T. Diab, Eneko Agirre, I{\~{n}}igo Lopez{-}Gazpio, and
  Lucia Specia. 2017.
\newblock \href {https://doi.org/10.18653/v1/S17-2001} {Semeval-2017 task 1:
  Semantic textual similarity multilingual and crosslingual focused
  evaluation}.
\newblock In \emph{Proceedings of the 11th International Workshop on Semantic
  Evaluation, SemEval@ACL 2017, Vancouver, Canada, August 3-4, 2017}, pages
  1--14.

\bibitem[{Chomsky(1980)}]{chomsky}
Noam Chomsky. 1980.
\newblock Rules and representations.
\newblock \emph{Behavioral and brain sciences}, 3(1):1--15.

\bibitem[{Chrupala et~al.(2015)Chrupala, K{\'{a}}d{\'{a}}r, and
  Alishahi}]{DBLP:conf/acl/ChrupalaKA15}
Grzegorz Chrupala, {\'{A}}kos K{\'{a}}d{\'{a}}r, and Afra Alishahi. 2015.
\newblock \href {http://aclweb.org/anthology/P/P15/P15-2019.pdf} {Learning
  language through pictures}.
\newblock In \emph{Proceedings of the 53rd Annual Meeting of the Association
  for Computational Linguistics and the 7th International Joint Conference on
  Natural Language Processing of the Asian Federation of Natural Language
  Processing, {ACL} 2015, July 26-31, 2015, Beijing, China, Volume 2: Short
  Papers}, pages 112--118.

\bibitem[{Collell and Moens(2018)}]{DBLP:conf/acl/CollellM18}
Guillem Collell and Marie{-}Francine Moens. 2018.
\newblock \href {https://aclanthology.info/papers/P18-2074/p18-2074} {Do neural
  network cross-modal mappings really bridge modalities?}
\newblock In \emph{Proceedings of the 56th Annual Meeting of the Association
  for Computational Linguistics, {ACL} 2018, Melbourne, Australia, July 15-20,
  2018, Volume 2: Short Papers}, pages 462--468.

\bibitem[{Collell et~al.(2017)Collell, Zhang, and Moens}]{guillem}
Guillem Collell, Teddy Zhang, and Marie-Francine Moens. 2017.
\newblock Imagined visual representations as multimodal embeddings.
\newblock In \emph{Proceedings of the Thirty-First AAAI Conference on
  Artificial Intelligence (AAAI-17)}. AAAI.

\bibitem[{Conneau and Kiela(2018)}]{conneau2018senteval}
Alexis Conneau and Douwe Kiela. 2018.
\newblock Senteval: An evaluation toolkit for universal sentence
  representations.
\newblock In \emph{Proceedings of the Eleventh International Conference on
  Language Resources and Evaluation, {LREC} 2018, Miyazaki, Japan, May 7-12,
  2018.}

\bibitem[{Conneau et~al.(2017)Conneau, Kiela, Schwenk, Barrault, and
  Bordes}]{DBLP:conf/emnlp/ConneauKSBB17}
Alexis Conneau, Douwe Kiela, Holger Schwenk, Lo{\"{\i}}c Barrault, and Antoine
  Bordes. 2017.
\newblock \href {https://aclanthology.info/papers/D17-1070/d17-1070}
  {Supervised learning of universal sentence representations from natural
  language inference data}.
\newblock In \emph{Proceedings of the 2017 Conference on Empirical Methods in
  Natural Language Processing, {EMNLP} 2017, Copenhagen, Denmark, September
  9-11, 2017}, pages 670--680.

\bibitem[{Dolan et~al.(2004)Dolan, Quirk, and Brockett}]{MSRP}
Bill Dolan, Chris Quirk, and Chris Brockett. 2004.
\newblock \href {http://www.aclweb.org/anthology/C04-1051} {Unsupervised
  construction of large paraphrase corpora: Exploiting massively parallel news
  sources}.
\newblock In \emph{{COLING} 2004, 20th International Conference on
  Computational Linguistics, Proceedings of the Conference, 23-27 August 2004,
  Geneva, Switzerland}.

\bibitem[{Fincher-Kiefer(2001)}]{Fincher-Kiefer2001}
Rebecca Fincher-Kiefer. 2001.
\newblock \href {https://doi.org/10.3758/BF03194928} {Perceptual components of
  situation models}.
\newblock \emph{Memory {\&} Cognition}, 29(2):336--343.

\bibitem[{Gordon and Van~Durme(2013)}]{Gordon:2013:RBK:2509558.2509563}
Jonathan Gordon and Benjamin Van~Durme. 2013.
\newblock \href {https://doi.org/10.1145/2509558.2509563} {Reporting bias and
  knowledge acquisition}.
\newblock In \emph{Proceedings of the 2013 Workshop on Automated Knowledge Base
  Construction}, AKBC '13, pages 25--30, New York, NY, USA. ACM.

\bibitem[{Harris(1954)}]{harris}
Zellig~S Harris. 1954.
\newblock Distributional structure.
\newblock \emph{Word}, 10(2-3):146--162.

\bibitem[{Hill et~al.(2016)Hill, Cho, and Korhonen}]{fastsent}
Felix Hill, Kyunghyun Cho, and Anna Korhonen. 2016.
\newblock \href {http://aclweb.org/anthology/N/N16/N16-1162.pdf} {Learning
  distributed representations of sentences from unlabelled data}.
\newblock In \emph{{NAACL} {HLT} 2016, The 2016 Conference of the North
  American Chapter of the Association for Computational Linguistics: Human
  Language Technologies, San Diego California, USA, June 12-17, 2016}, pages
  1367--1377.

\bibitem[{Hill and Korhonen(2014)}]{DBLP:conf/emnlp/HillK14}
Felix Hill and Anna Korhonen. 2014.
\newblock \href {http://aclweb.org/anthology/D/D14/D14-1032.pdf} {Learning
  abstract concept embeddings from multi-modal data: Since you probably can't
  see what {I} mean}.
\newblock In \emph{Proceedings of the 2014 Conference on Empirical Methods in
  Natural Language Processing, {EMNLP} 2014, October 25-29, 2014, Doha, Qatar,
  {A} meeting of SIGDAT, a Special Interest Group of the {ACL}}, pages
  255--265.

\bibitem[{Hill et~al.(2014)Hill, Reichart, and Korhonen}]{weighted-gram}
Felix Hill, Roi Reichart, and Anna Korhonen. 2014.
\newblock \href {http://aclweb.org/anthology/Q14-1023} {Multi-modal models for
  concrete and abstract concept meaning}.
\newblock \emph{Transactions of the Association for Computational Linguistics},
  2:285--296.

\bibitem[{Hu and Liu(2004)}]{CR}
Minqing Hu and Bing Liu. 2004.
\newblock \href {https://doi.org/10.1145/1014052.1014073} {Mining and
  summarizing customer reviews}.
\newblock In \emph{Proceedings of the Tenth ACM SIGKDD International Conference
  on Knowledge Discovery and Data Mining}, KDD '04, pages 168--177, New York,
  NY, USA. ACM.

\bibitem[{Kalchbrenner et~al.(2014)Kalchbrenner, Grefenstette, and
  Blunsom}]{DBLP:conf/acl/KalchbrennerGB14}
Nal Kalchbrenner, Edward Grefenstette, and Phil Blunsom. 2014.
\newblock \href {http://aclweb.org/anthology/P/P14/P14-1062.pdf} {A
  convolutional neural network for modelling sentences}.
\newblock In \emph{Proceedings of the 52nd Annual Meeting of the Association
  for Computational Linguistics, {ACL} 2014, June 22-27, 2014, Baltimore, MD,
  USA, Volume 1: Long Papers}, pages 655--665.

\bibitem[{Karpathy and Li(2015)}]{DBLP:conf/cvpr/KarpathyL15}
Andrej Karpathy and Fei{-}Fei Li. 2015.
\newblock \href {https://doi.org/10.1109/CVPR.2015.7298932} {Deep
  visual-semantic alignments for generating image descriptions}.
\newblock In \emph{{IEEE} Conference on Computer Vision and Pattern
  Recognition, {CVPR} 2015, Boston, MA, USA, June 7-12, 2015}, pages
  3128--3137.

\bibitem[{Kiela et~al.(2018)Kiela, Conneau, Jabri, and
  Nickel}]{DBLP:conf/naacl/KielaCJN18}
Douwe Kiela, Alexis Conneau, Allan Jabri, and Maximilian Nickel. 2018.
\newblock \href {https://aclanthology.info/papers/N18-1038/n18-1038} {Learning
  visually grounded sentence representations}.
\newblock In \emph{Proceedings of the 2018 Conference of the North American
  Chapter of the Association for Computational Linguistics: Human Language
  Technologies, {NAACL-HLT} 2018, New Orleans, Louisiana, USA, June 1-6, 2018,
  Volume 1 (Long Papers)}, pages 408--418.

\bibitem[{Kingma and Ba(2014)}]{DBLP:journals/corr/KingmaB14}
Diederik~P. Kingma and Jimmy Ba. 2014.
\newblock \href {http://arxiv.org/abs/1412.6980} {Adam: {A} method for
  stochastic optimization}.
\newblock \emph{CoRR}, abs/1412.6980.

\bibitem[{Kiros et~al.(2015)Kiros, Zhu, Salakhutdinov, Zemel, Urtasun,
  Torralba, and Fidler}]{skip_thought}
Ryan Kiros, Yukun Zhu, Ruslan Salakhutdinov, Richard~S. Zemel, Raquel Urtasun,
  Antonio Torralba, and Sanja Fidler. 2015.
\newblock \href {http://papers.nips.cc/paper/5950-skip-thought-vectors}
  {Skip-thought vectors}.
\newblock In \emph{Advances in Neural Information Processing Systems 28: Annual
  Conference on Neural Information Processing Systems 2015, December 7-12,
  2015, Montreal, Quebec, Canada}, pages 3294--3302.

\bibitem[{Kottur et~al.(2016)Kottur, Vedantam, Moura, and
  Parikh}]{DBLP:conf/cvpr/KotturVMP16}
Satwik Kottur, Ramakrishna Vedantam, Jos{\'{e}} M.~F. Moura, and Devi Parikh.
  2016.
\newblock \href {https://doi.org/10.1109/CVPR.2016.539} {Visualword2vec
  (vis-w2v): Learning visually grounded word embeddings using abstract scenes}.
\newblock In \emph{2016 {IEEE} Conference on Computer Vision and Pattern
  Recognition, {CVPR} 2016, Las Vegas, NV, USA, June 27-30, 2016}, pages
  4985--4994.

\bibitem[{Lazaridou et~al.(2015{\natexlab{a}})Lazaridou, Dinu, and
  Baroni}]{lazaridou}
Angeliki Lazaridou, Georgiana Dinu, and Marco Baroni. 2015{\natexlab{a}}.
\newblock \href {https://doi.org/10.3115/v1/P15-1027} {Hubness and pollution:
  Delving into cross-space mapping for zero-shot learning}.
\newblock In \emph{Proceedings of the 53rd Annual Meeting of the Association
  for Computational Linguistics and the 7th International Joint Conference on
  Natural Language Processing (Volume 1: Long Papers)}, pages 270--280.
  Association for Computational Linguistics.

\bibitem[{Lazaridou et~al.(2015{\natexlab{b}})Lazaridou, Pham, and
  Baroni}]{Multimodal_Skipgram}
Angeliki Lazaridou, Nghia~The Pham, and Marco Baroni. 2015{\natexlab{b}}.
\newblock \href {http://aclweb.org/anthology/N/N15/N15-1016.pdf} {Combining
  language and vision with a multimodal skip-gram model}.
\newblock In \emph{{NAACL} {HLT} 2015, The 2015 Conference of the North
  American Chapter of the Association for Computational Linguistics: Human
  Language Technologies, Denver, Colorado, USA, May 31 - June 5, 2015}, pages
  153--163.

\bibitem[{Lin et~al.(2014)Lin, Maire, Belongie, Hays, Perona, Ramanan,
  Doll{\'{a}}r, and Zitnick}]{mscoco}
Tsung{-}Yi Lin, Michael Maire, Serge~J. Belongie, James Hays, Pietro Perona,
  Deva Ramanan, Piotr Doll{\'{a}}r, and C.~Lawrence Zitnick. 2014.
\newblock \href {https://doi.org/10.1007/978-3-319-10602-1\_48} {Microsoft
  {COCO:} common objects in context}.
\newblock In \emph{Computer Vision - {ECCV} 2014 - 13th European Conference,
  Zurich, Switzerland, September 6-12, 2014, Proceedings, Part {V}}, pages
  740--755.

\bibitem[{Lin and Parikh(2015)}]{DBLP:conf/cvpr/LinP15}
Xiao Lin and Devi Parikh. 2015.
\newblock \href {https://doi.org/10.1109/CVPR.2015.7298917} {Don't just listen,
  use your imagination: Leveraging visual common sense for non-visual tasks}.
\newblock In \emph{{IEEE} Conference on Computer Vision and Pattern
  Recognition, {CVPR} 2015, Boston, MA, USA, June 7-12, 2015}, pages
  2984--2993.

\bibitem[{Lin et~al.(2017)Lin, Feng, dos Santos, Yu, Xiang, Zhou, and
  Bengio}]{Attentive}
Zhouhan Lin, Minwei Feng, C{\'{\i}}cero~Nogueira dos Santos, Mo~Yu, Bing Xiang,
  Bowen Zhou, and Yoshua Bengio. 2017.
\newblock \href {http://arxiv.org/abs/1703.03130} {A structured self-attentive
  sentence embedding}.
\newblock \emph{CoRR}, abs/1703.03130.

\bibitem[{Logeswaran and Lee(2018)}]{quickthought}
Lajanugen Logeswaran and Honglak Lee. 2018.
\newblock \href {https://openreview.net/forum?id=rJvJXZb0W} {An efficient
  framework for learning sentence representations}.
\newblock In \emph{6th International Conference on Learning Representations,
  {ICLR} 2018, Vancouver, BC, Canada, April 30 - May 3, 2018, Conference Track
  Proceedings}.

\bibitem[{Maaten and Hinton(2008)}]{maaten2008visualizing}
Laurens van~der Maaten and Geoffrey Hinton. 2008.
\newblock Visualizing data using t-sne.
\newblock \emph{Journal of machine learning research}, 9(Nov):2579--2605.

\bibitem[{Marelli et~al.(2014{\natexlab{a}})Marelli, Bentivogli, Baroni,
  Bernardi, Menini, and Zamparelli}]{SICK}
Marco Marelli, Luisa Bentivogli, Marco Baroni, Raffaella Bernardi, Stefano
  Menini, and Roberto Zamparelli. 2014{\natexlab{a}}.
\newblock \href {http://aclweb.org/anthology/S/S14/S14-2001.pdf} {Semeval-2014
  task 1: Evaluation of compositional distributional semantic models on full
  sentences through semantic relatedness and textual entailment}.
\newblock In \emph{Proceedings of the 8th International Workshop on Semantic
  Evaluation, SemEval@COLING 2014, Dublin, Ireland, August 23-24, 2014.}, pages
  1--8.

\bibitem[{Marelli et~al.(2014{\natexlab{b}})Marelli, Menini, Baroni,
  Bentivogli, Bernardi, and Zamparelli}]{DBLP:conf/lrec/MarelliMBBBZ14}
Marco Marelli, Stefano Menini, Marco Baroni, Luisa Bentivogli, Raffaella
  Bernardi, and Roberto Zamparelli. 2014{\natexlab{b}}.
\newblock \href
  {http://www.lrec-conf.org/proceedings/lrec2014/summaries/363.html} {A {SICK}
  cure for the evaluation of compositional distributional semantic models}.
\newblock In \emph{Proceedings of the Ninth International Conference on
  Language Resources and Evaluation, {LREC} 2014, Reykjavik, Iceland, May
  26-31, 2014.}, pages 216--223.

\bibitem[{Mikolov et~al.(2013)Mikolov, Sutskever, Chen, Corrado, and
  Dean}]{mikolov}
Tomas Mikolov, Ilya Sutskever, Kai Chen, Gregory~S. Corrado, and Jeffrey Dean.
  2013.
\newblock \href
  {http://papers.nips.cc/paper/5021-distributed-representations-of-words-and-phrases-and-their-compositionality}
  {Distributed representations of words and phrases and their
  compositionality}.
\newblock In \emph{Advances in Neural Information Processing Systems 26: 27th
  Annual Conference on Neural Information Processing Systems 2013. Proceedings
  of a meeting held December 5-8, 2013, Lake Tahoe, Nevada, United States.},
  pages 3111--3119.

\bibitem[{Nelson et~al.(2004)Nelson, McEvoy, and Schreiber}]{usf}
Douglas~L Nelson, Cathy~L McEvoy, and Thomas~A Schreiber. 2004.
\newblock The university of south florida free association, rhyme, and word
  fragment norms.
\newblock \emph{Behavior Research Methods, Instruments, \& Computers},
  36(3):402--407.

\bibitem[{Norman(1972)}]{norman1972memory}
Donald~A Norman. 1972.
\newblock Memory, knowledge, and the answering of questions.

\bibitem[{Pang and Lee(2004)}]{SUBJ}
Bo~Pang and Lillian Lee. 2004.
\newblock \href {http://aclweb.org/anthology/P/P04/P04-1035.pdf} {A sentimental
  education: Sentiment analysis using subjectivity summarization based on
  minimum cuts}.
\newblock In \emph{Proceedings of the 42nd Annual Meeting of the Association
  for Computational Linguistics, 21-26 July, 2004, Barcelona, Spain.}, pages
  271--278.

\bibitem[{Pang and Lee(2005)}]{MR}
Bo~Pang and Lillian Lee. 2005.
\newblock \href {http://aclweb.org/anthology/P/P05/P05-1015.pdf} {Seeing stars:
  Exploiting class relationships for sentiment categorization with respect to
  rating scales}.
\newblock In \emph{{ACL} 2005, 43rd Annual Meeting of the Association for
  Computational Linguistics, Proceedings of the Conference, 25-30 June 2005,
  University of Michigan, {USA}}, pages 115--124.

\bibitem[{Peters et~al.(2018)Peters, Neumann, Iyyer, Gardner, Clark, Lee, and
  Zettlemoyer}]{elmo}
Matthew~E. Peters, Mark Neumann, Mohit Iyyer, Matt Gardner, Christopher Clark,
  Kenton Lee, and Luke Zettlemoyer. 2018.
\newblock \href {https://aclanthology.info/papers/N18-1202/n18-1202} {Deep
  contextualized word representations}.
\newblock In \emph{Proceedings of the 2018 Conference of the North American
  Chapter of the Association for Computational Linguistics: Human Language
  Technologies, {NAACL-HLT} 2018, New Orleans, Louisiana, USA, June 1-6, 2018,
  Volume 1 (Long Papers)}, pages 2227--2237.

\bibitem[{Plummer et~al.(2015)Plummer, Wang, Cervantes, Caicedo, Hockenmaier,
  and Lazebnik}]{DBLP:conf/iccv/PlummerWCCHL15}
Bryan~A. Plummer, Liwei Wang, Chris~M. Cervantes, Juan~C. Caicedo, Julia
  Hockenmaier, and Svetlana Lazebnik. 2015.
\newblock \href {https://doi.org/10.1109/ICCV.2015.303} {Flickr30k entities:
  Collecting region-to-phrase correspondences for richer image-to-sentence
  models}.
\newblock In \emph{2015 {IEEE} International Conference on Computer Vision,
  {ICCV} 2015, Santiago, Chile, December 7-13, 2015}, pages 2641--2649.

\bibitem[{Silberer and Lapata(2014)}]{autoencoders}
Carina Silberer and Mirella Lapata. 2014.
\newblock \href {http://aclweb.org/anthology/P/P14/P14-1068.pdf} {Learning
  grounded meaning representations with autoencoders}.
\newblock In \emph{Proceedings of the 52nd Annual Meeting of the Association
  for Computational Linguistics, {ACL} 2014, June 22-27, 2014, Baltimore, MD,
  USA, Volume 1: Long Papers}, pages 721--732.

\bibitem[{Socher et~al.(2013)Socher, Perelygin, Wu, Chuang, Manning, Ng, and
  Potts}]{DBLP:conf/emnlp/SocherPWCMNP13}
Richard Socher, Alex Perelygin, Jean Wu, Jason Chuang, Christopher~D. Manning,
  Andrew~Y. Ng, and Christopher Potts. 2013.
\newblock \href {https://aclanthology.info/papers/D13-1170/d13-1170} {Recursive
  deep models for semantic compositionality over a sentiment treebank}.
\newblock In \emph{Proceedings of the 2013 Conference on Empirical Methods in
  Natural Language Processing, {EMNLP} 2013, 18-21 October 2013, Grand Hyatt
  Seattle, Seattle, Washington, USA, {A} meeting of SIGDAT, a Special Interest
  Group of the {ACL}}, pages 1631--1642.

\bibitem[{Szegedy et~al.(2016)Szegedy, Vanhoucke, Ioffe, Shlens, and
  Wojna}]{DBLP:conf/cvpr/SzegedyVISW16}
Christian Szegedy, Vincent Vanhoucke, Sergey Ioffe, Jonathon Shlens, and
  Zbigniew Wojna. 2016.
\newblock \href {https://doi.org/10.1109/CVPR.2016.308} {Rethinking the
  inception architecture for computer vision}.
\newblock In \emph{2016 {IEEE} Conference on Computer Vision and Pattern
  Recognition, {CVPR} 2016, Las Vegas, NV, USA, June 27-30, 2016}, pages
  2818--2826.

\bibitem[{W.~Barsalou(1999)}]{barsalou}
Lawrence W.~Barsalou. 1999.
\newblock Perceptual symbol systems.
\newblock 22:577--609; discussion 610.

\bibitem[{Wang et~al.(2019)Wang, Xiong, Yu, Guo, Chang, and
  Wang}]{relationextraction}
Hong Wang, Wenhan Xiong, Mo~Yu, Xiaoxiao Guo, Shiyu Chang, and William~Yang
  Wang. 2019.
\newblock \href {http://arxiv.org/abs/1903.02588} {Sentence embedding alignment
  for lifelong relation extraction}.
\newblock \emph{NAACL}.

\bibitem[{Wang et~al.(2016)Wang, Li, and Lazebnik}]{DBLP:conf/cvpr/WangLL16}
Liwei Wang, Yin Li, and Svetlana Lazebnik. 2016.
\newblock \href {https://doi.org/10.1109/CVPR.2016.541} {Learning deep
  structure-preserving image-text embeddings}.
\newblock In \emph{2016 {IEEE} Conference on Computer Vision and Pattern
  Recognition, {CVPR} 2016, Las Vegas, NV, USA, June 27-30, 2016}, pages
  5005--5013.

\bibitem[{Wiebe and Cardie(2005)}]{MPQA}
Janyce Wiebe and Claire Cardie. 2005.
\newblock Annotating expressions of opinions and emotions in language. language
  resources and evaluation.
\newblock In \emph{Language Resources and Evaluation (formerly Computers and
  the Humanities}, page 2005.

\bibitem[{Xiao et~al.(2017)Xiao, Sigal, and Lee}]{DBLP:conf/cvpr/XiaoSL17}
Fanyi Xiao, Leonid Sigal, and Yong~Jae Lee. 2017.
\newblock \href {https://doi.org/10.1109/CVPR.2017.558} {Weakly-supervised
  visual grounding of phrases with linguistic structures}.
\newblock In \emph{2017 {IEEE} Conference on Computer Vision and Pattern
  Recognition, {CVPR} 2017, Honolulu, HI, USA, July 21-26, 2017}, pages
  5253--5262.

\bibitem[{Yatskar et~al.(2016)Yatskar, Ordonez, and
  Farhadi}]{DBLP:conf/naacl/YatskarOF16}
Mark Yatskar, Vicente Ordonez, and Ali Farhadi. 2016.
\newblock \href {http://aclweb.org/anthology/N/N16/N16-1023.pdf} {Stating the
  obvious: Extracting visual common sense knowledge}.
\newblock In \emph{{NAACL} {HLT} 2016, The 2016 Conference of the North
  American Chapter of the Association for Computational Linguistics: Human
  Language Technologies, San Diego California, USA, June 12-17, 2016}, pages
  193--198.

\bibitem[{Zablocki et~al.(2018)Zablocki, Piwowarski, Soulier, and
  Gallinari}]{zablocki}
\'Eloi Zablocki, Benjamin Piwowarski, Laure Soulier, and Patrick Gallinari.
  2018.
\newblock \href
  {https://www.aaai.org/ocs/index.php/AAAI/AAAI18/paper/view/16113} {Learning
  multi-modal word representation grounded in visual context}.
\newblock In \emph{Proceedings of the Thirty-Second {AAAI} Conference on
  Artificial Intelligence, New Orleans, Louisiana, USA, February 2-7, 2018}.

\end{thebibliography}
\bibliographystyle{acl_natbib}
\end{document}